\documentclass[final,3p,times]{elsarticle}

\usepackage{amssymb}
\usepackage{amsthm}

\usepackage{multirow}
\usepackage{subcaption}
\usepackage{graphicx}
\usepackage{booktabs}
\usepackage{xcolor}
\usepackage{url}
\usepackage{hyperref}
\usepackage{pifont}%
\usepackage{amsmath}

\newcommand{\cmark}{\color{blue!20!red!20!green}\ding{51}}%
\newcommand{\xmark}{\color{red!75}\ding{55}}%

\journal{Knowledge-Based Systems}
\begin{document}

\begin{frontmatter}

\title{Prompting Large Language Models with Knowledge Graphs for Question Answering Involving Long-tail Facts}

\author[uoe]{Wenyu Huang}
\author[huawei]{Guancheng Zhou}
\author[uoe]{Mirella Lapata}
\author[huawei]{Pavlos Vougiouklis}
\author[huawei]{Sebastien Montella}
\author[uoe,huawei]{Jeff Z. Pan}

\affiliation[uoe]{organization={School of Informatics, University of Edinburgh},%
            city={Edinburgh},
            country={UK}}
            
\affiliation[huawei]{organization={Huawei Edinburgh Research Centre, CSI},%
            city={Edinburgh},
            country={UK}}

\begin{abstract}
Although Large Language Models (LLMs) are effective in performing various NLP tasks, they still struggle to handle tasks that require extensive, real-world knowledge, especially when dealing with long-tail facts (facts related to long-tail entities). This limitation highlights the need to supplement LLMs with non-parametric knowledge. To address this issue, we analysed the effects of different types of non-parametric knowledge, including textual passage and knowledge graphs (KGs). Since LLMs have probably seen the majority of factual question-answering datasets already, to facilitate our analysis, we proposed a fully automatic pipeline for creating a benchmark that requires knowledge of long-tail facts for answering the involved questions. Using this pipeline, we introduce the LTGen benchmark.
We evaluate state-of-the-art LLMs in different knowledge settings using the proposed benchmark.
Our experiments show that LLMs alone struggle with answering these questions, especially when the long-tail level is high or rich knowledge is required. Nonetheless, the performance of the same models improved significantly when they were prompted with non-parametric knowledge.
We observed that, in most cases, prompting LLMs with KG triples surpasses passage-based prompting using a state-of-the-art retriever. In addition, while prompting LLMs with both KG triples and documents does not consistently improve knowledge coverage, it can dramatically reduce hallucinations in the generated content.

\end{abstract}


\end{frontmatter}

\section{Introduction}
Large Language Models (LLMs), such as GPT-4 \citep{DBLP:journals/corr/abs-2303-08774}, LLaMA \citep{DBLP:journals/corr/abs-2302-13971, DBLP:journals/corr/abs-2307-09288} and PaLM2 \citep{DBLP:journals/corr/abs-2305-10403}, have shown impressive ability in conversational search. While these LLMs can generate knowledgeable responses, they can still suffer from hallucinations in Knowledge-Intensive Generation (KIG) tasks \citep{petroni-etal-2021-kilt}, in particular for cases involving long-tail entities.

Prompting LLMs with non-parametric knowledge has been recently proposed as a method for mitigating this issue \citep{mallen-etal-2023-trust, DBLP:journals/corr/abs-2302-12813}.
\citet{mallen-etal-2023-trust} investigated the effectiveness of prompting LLMs with non-parametric knowledge, showing 
its effectiveness when coupled with the parameterized knowledge of LLMs for long-tail question answering.
Their investigation was conducted on the PopQA dataset, which is constructed from Knowledge Graph (KG)~\cite{PVGW2017,Pan2017b} triples using predefined templates. While their study exhibited promising  results for motivating further research in this space, we believe it was also subjected to certain \emph{limitations}: 1) the template-based dataset creation-process limits the predicates (relations) of the selected triples to be investigated; 2) each question in PopQA is constructed from a single entity-relation pair; 
3) the authors only considered unstructured, non-parametric knowledge, while ignoring any type of structured, non-parametric knowledge, for example the one included in KGs.

\begin{figure}
    \centering
    \includegraphics[width=0.96\textwidth]{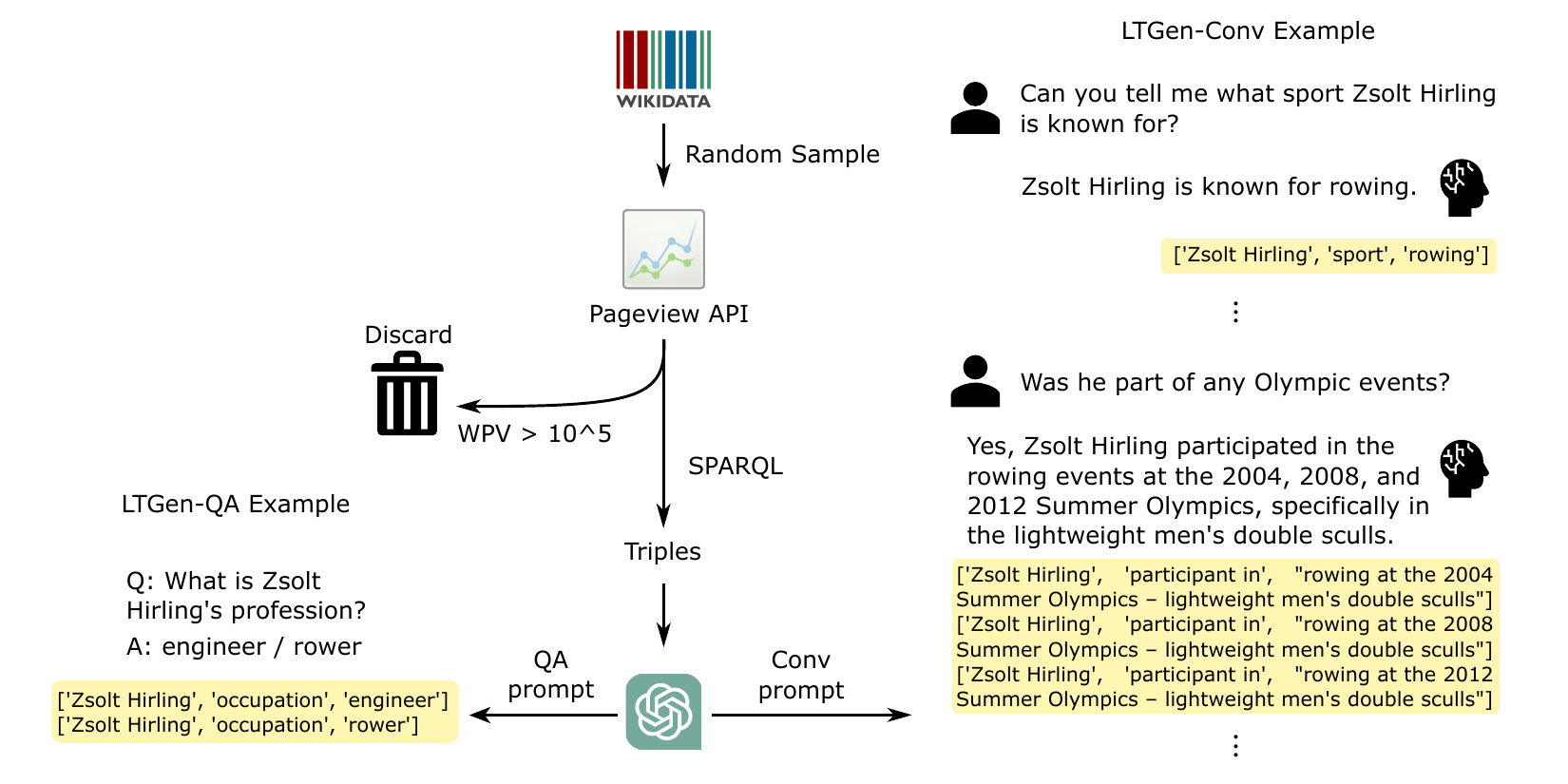}
    \caption{Pipeline overview for generating LTGen-GPT dataset. In the implementation, we use \texttt{gpt-4-turbo} to generate dialogues. \colorbox{yellow!40}{Gold highlighted} triples are gold triples selected by GPT for use as the external knowledge source to respond current query.}
    \label{fig:ltgen-gpt}
\end{figure}

In this work, we propose a fully automatic, template-free pipeline for building long-tail generation benchmarks.
Using the proposed pipeline (see Figure~\ref{fig:ltgen-gpt}), we introduce the Long Tail Generation (LTGen) benchmark. This benchmark comprises two tasks: question answering (LTGen-QA) and conversational QA (LTGen-Conv).
Compared to PopQA, LTGen: (1) does not rely on templates; (2) contains questions constructed from multiple relations (in LTGen-Conv); and (3) provides a more challenging task of conversational QA (LTGen-Conv).
The key research questions that motivate this work are as follows:

\begin{itemize}
    \item \textbf{RQ1}: How do LLMs perform in knowledge-intensive generation tasks that involve different kinds of factual knowledge?
    \item \textbf{RQ3}: How LLMs benefit from different formats of non-parametric knowledge (structured and unstructured) with different kinds of factual knowledge?
\end{itemize}

Our evaluation demonstrates that LLMs equipped solely with their parametric knowledge exhibit poor performance when tasked with answering questions involving long-tail facts.  Their performance declines as both the long-tail level of the relevant entities and the required amount of knowledge (number of reference triples) increases.
Prompting LLMs with non-parametric knowledge can significantly benefit performance in long-tail question-answering. Such benefits are even more evident in the case of small language models \citep{mallen-etal-2023-trust}. 
Furthermore, our results show that KG triples can serve as a powerful and robust source of non-parametric knowledge while offering a more efficient processing paradigm than the unstructured nature of textual passages. Moreover, the amalgamation of both structural knowledge from KGs and unstructured knowledge from passages can reduce hallucination tendencies of the involved LLMs, leading to the highest NLI-based metric scores across various settings.

\section{Related Works}
\paragraph{Knowledge Bases (KBs)}
Typical KBs, such as Wikidata \citep{DBLP:journals/cacm/VrandecicK14}, comprise several key elements. The   basic unit is   entity, which forms the cornerstone of a KB. Subsequently, concepts and attributes come completing the details of   entities:
the former are abstractions representing a group of entities, whereas the latter provide literal and descriptive information about a specific entity where each attribute is defined by a key and a value. Furthermore, relations are a \textit{sine qua non} component of KB as they establish the relationships between entities or concepts. For instance, entities are connected to concepts via the 'instance of' relation, and concepts are systematically arranged into a hierarchical tree through the 'subclass of' relation.

The knowledge within a KB is encapsulated in three forms: %
\begin{enumerate}
    \item Relational Knowledge: Structured as triples where each triple consists of two entities linked by a relation.
    \item Literal Knowledge: Structured as triples where an entity is associated with an attribute key and an attribute value
    \item Qualifier Knowledge extends these basic triples, whose head is a relational or literal triple. A qualifier also has a key and value \citep{cao-etal-2022-kqa,HBXLP2023}.
\end{enumerate}
 
\paragraph{Question Answering Over Knowledge Bases (KBQA)}
KBQA is a long-studied classical NLP task and has gained widespread application in recent times due to large language models (LLMs)~\cite{HWQM+2023}. Unlike approaches that depend on parametric knowledge, KBQA leverages structured knowledge bases to derive answers. Complex KBQA systems can be broadly classified into two categories: Semantic Parsing (SP)-based~\cite{conf/eacl/PerezBeltrachiniJML23} and Information Retrieval (IR)-based techniques \citep{DBLP:conf/ijcai/LanHJ0ZW21}. Each category employs distinct mechanisms to tackle the KBQA task. SP-based methods convert a question into a symbolic logical form, which is then executed on the KB to extract precise answers. In contrast, IR-based approaches focus on gathering comprehensive information related to the query, ranking entities and relations by their relevance to the posed question. However, an increasing trend in the field is moving away from the traditional single-turn KBQA approach towards a more conversational setting due to the notable shift in the industrial applications~\cite{journals/kais/ZaibZSMZ22}.

\paragraph{Understanding Complex Semantics and Syntax}
Complex questions are distinguished by their complex query types and compositional semantics. This complexity presents significant challenges in linguistic analysis. To make complex problems clearer, it is good practice to convert questions from their unstructured forms (i.e. natural text) to a more structural or logical representation, such as SQL~\cite{VPZT+2023,ZLP2024}. To bridge natural language questions with their corresponding logical expressions, methodologies like NSQA \citep{kapanipathi-etal-2021-leveraging} adopt Abstract Meaning Representation (AMR) as an intermediary logical form. This approach aligns the elements of a question, such as entities, relations, entity types, and attributes, with their syntactic counterparts, enhancing the coherence between these components and syntax. The NSQA model initially parsed the question into a rooted, directed, and acyclic AMR graph. Subsequently, it involved entity linking within the schema ground phase, aligning the entity nodes in the AMR graph with KBs. Building on these aligned entities, NSQA employs a path-expanding methodology to convert the AMR graph into a predefined query graph. This graph could then be converted into executable logical expressions easily \citep{DBLP:conf/ijcai/LanHJ0ZW21, ZHANG20231}.
\paragraph{KIG Benchmarks}
Traditional KIG benchmarks as KILT \citep{petroni-etal-2021-kilt} do not especially focus on the long-tail problem, thus having very few long-tail examples in their corresponding datasets. 
Table~\ref{tab:kilt_res} in \ref{app:kilt} reveals the performance of GPT3.5 and RE2G (i.e. a state-of-the-art retrieval-augmented generation method) \citep{glass-etal-2022-re2g} in NQ and TQA benchmarks. While being totally devoid of any non-parametric memory such as KG, GPT3.5 outperforms RE2G on the TQA dataset by a significant margin. However, it %
remains unclear whether LLMs actually encode the factual knowledge~\cite{PRKS+2023} required to answer TQA questions, %
or simply memorize the question-answer pairs in the datasets \cite{petroni-etal-2019-language}. %
We remark importantly that the recent LLMs (which are continuously upgraded) easily outmatch the state-of-the-art on initially released datasets.
This strengthens the need of challenging benchmarks for robust, fair and ethical evaluation of emerging LLMs.

\paragraph{Long-tail benchmarks}
PopQA \citep{mallen-etal-2023-trust} is a benchmark used to assess the performance of language models in addressing queries pertaining to long-tail entities. The questions are generated from triples with predefined templates for a fixed set of relations. Limited by these predefined templates, PopQA only contains limited relations. This drawback limits PopQA's ability to evaluate LLMs for diverse relation types that today's users would ask.

\section{LTGen: Long-Tail Generation Tasks}

To address the aforementioned limitations, we introduce a novel pipeline to automatically generate a high-quality KIG dataset. We further release the resulting \textbf{L}ong-\textbf{T}ail \textbf{Gen}eration (LTGen) benchmark which includes two datasets: LTGen-QA and LTGen-Conv. The former gathers simple question answering samples while the latter involves multiple relations in a multi-turn scenario. Table~\ref{tab:dataset_compare} compares our benchmark to other relevant works, showing its completeness across a set of different criteria.

\begin{table}[t]
    \centering
    \small
    \begin{tabular}{ccccc}
    \toprule
    Dataset & Long-Tail & Conversation & KG Triples & Template Free \\ \midrule
    Natural Questions \citep{kwiatkowski-etal-2019-natural} & \xmark & \xmark & \xmark & - \\
    TriviaQA \citep{joshi-etal-2017-triviaqa} & \xmark & \xmark & \xmark & - \\ 
    MSMARCO \citep{DBLP:conf/nips/NguyenRSGTMD16} & \xmark & \xmark & \xmark & - \\ 
    Wizard of Wikipedia \citep{DBLP:conf/iclr/DinanRSFAW19} & \xmark & \cmark & \xmark & - \\ \midrule
    PopQA \citep{mallen-etal-2023-trust} & \cmark & \xmark & \cmark & \xmark \\
    \midrule
    LTGen-QA & \cmark & \xmark & \cmark & \cmark \\
    LTGen-Conv & \cmark & \cmark & \cmark & \cmark \\
    \bottomrule
    \end{tabular}
    \caption{Compare LTGen benchmark with related works.}
    \label{tab:dataset_compare}
\end{table}

\subsection{Benchmark Construction}
\label{sec:benchmark_construction}
We propose an automatic data generation pipeline with the help of LLMs. This is inspired by many recent works that have looked into using LLMs to generate high-quality datasets for training smaller models \citep{DBLP:journals/corr/abs-2212-10560, alpaca, DBLP:journals/corr/abs-2304-03277}. The pipeline is composed of three stages: Long-tail entity selection, triples retrieval and sample generation. We hereinafter detail the process for each step.

\begin{table}[t]
    \centering
    \small
    \begin{tabular}{cccccc}
    \toprule
        Long-tail Level & WPV range & Entities & Relations & Samples & avg. Turns \\ \midrule
        \multicolumn{6}{c}{\textit{LTGen-QA}} \\ \midrule
        Overall & ($10^1$, $10^5$) & 2,000 & 439 & 6,000   & - \\
        I & ($10^4$, $10^5$)& 500 & 291 & 1,500 & -\\
        II & ($10^3$, $10^4$) & 500 & 230 & 1,500 & -\\
        III & ($10^2$, $10^3$) & 500 & 215 & 1,500 & -\\
        IV & ($10^1$, $10^2$) & 500 & 207 & 1,500 & -\\
        \midrule
        \multicolumn{6}{c}{\textit{LTGen-Conv}} \\ \midrule
        Overall & ($10^1$, $10^5$) & 2,000 & 592 & 13,334 & 7.686\\
        I & ($10^4$, $10^5$)& 500 & 409 & 3,617 & 8.252 \\
        II & ($10^3$, $10^4$) & 500 & 345 & 3,308 & 7.636 \\
        III & ($10^2$, $10^3$) & 500 & 328 & 3,199 & 7.424 \\
        IV & ($10^1$, $10^2$) & 500 & 276 & 3,210 & 7.432 \\
    \bottomrule
    \end{tabular}
    \caption{Dataset Statistic of LTGen Benchmark}
    \label{tab:dataset_statistic}
\end{table}
\paragraph{Long-tail Entity Sampling}Determining whether an entity can be classified as long-tail is not only a complex task, but also highly subjective. To this end, we seek to establish a clear and universally-accepted definition which ensures a precise categorization of any involved entities. We follow the assumption by \citep{mallen-etal-2023-trust} which considers an entity as rare if its corresponding Wikipedia page is among the least visited ones. Thus, for each Wikidata entity, we use the Wikipedia Pageview API\footnote{\url{https://wikitech.wikimedia.org/wiki/Analytics/AQS/Pageviews}} to get the average monthly Wikipedia Page View (WPV) to determine the popularity of an entity. To generate the LTGen benchmark, we first randomly sample entities from Wikidata. Then, we select long-tail entities based on their popularity using their respective WPV\footnote{Average monthly Wikipedia page viewing count from 1 Jan 2021 to 1 Jan 2023.}.

\paragraph{Triples Retrieval}Using the sampled long-tail entities, we design SPARQL queries to retrieve relevant triples, i.e. triples where the entities appear either in the subject or the object position. Nevertheless, accumulating all of those triples may not be judicious because of the non-informative relations as ``family name'' or ``given name'' in Wikidata. As a result, to prevent the generation of trivial questions like ``What is the family name of Joe Biden?'', we remove triples with these non-informative relations.
Next, we filter out any entities appearing in less than 5 triples or more than 100 triples. Removing the former %
makes the task more challenging, while removing the latter due to  
consideration of the efficiency of prompting. %
\paragraph{Samples Generation}
The generation of a new dataset is usually carried out by leveraging predefined templates, LLM-based prompting or combinations thereof. With the recent breakthrough achieved in prompt engineering, we set our data generator as GPT-4. We use GPT-4 to create PopQA-like samples where each question is constructed from an entity-relation pair. In such a template-free setting, we can fascilitate the inclusion of more complex and diverse QA samples that consider more relation types. Moreover, we further prompt GPT-4 to generate dialogues from the collected triples to build LTGen-Conv. This enables us to annotate automatically a generated dialogue against gold triples that are relevant to its textual content. Figure~\ref{fig:ltgen-gpt} shows the overall workflow of generating the LTGen benchmark. The final prompts for LTGen-QA and LTGen-Conv construction are detailed in \ref{app:prompt}.

Table~\ref{tab:dataset_statistic} shows the overall statistics of the proposed LTGen benchmark. We additionally define four long-tail levels, identified as level I, II, III, and IV, which are based on the count of WPV, with the level IV being the highest degree of rarity.
Note that LTGen-Conv contains questions that require multiple relations to answer, %
allowing us to evaluate LLMs from that perspective. Therefore, we further split LTGen-Conv into four subsets with regard to the number of reference triples, which are 1) single reference triple with 10,330 samples; 2) two reference triples with 1,720 samples; 3) three reference triples with 590 samples and 4) more than three reference triples with 694 samples.

\subsection{Data Quality Checking}
\label{sec:data_quality}
To ensure data quality, we randomly select 3\% of the LTGen-Conv datasets standing for 400 samples\footnote{A relatively high rate compared to related works \citep{elsahar-etal-2018-rex}} and we comply with a thorough manual quality check procedure. More specifically, for each selected sample, we hire two annotators to review the associated triple annotations following the instructions detailed in \ref{app:data_quality}. We calculate the frequency of irrelevant triples linked to or missing triples required for a specific dialogue turn. Among the 400 data samples, we found an average of 9 data samples with incorrect annotations, representing an approximate 2.25\% error rate. Considering the in-annotator outcome, where correctness is determined only if both annotators concur, we identified 12 data samples, resulting in an approximately 3\% error rate. \ref{app:data_quality} provides more details of the data quality checking process.

\section{Non-parametric Memories Collection}
\label{sec:knowledge}
We consider two sources of non-parametric knowledge~\cite{PRKS+2023} which are (i) unstructured knowledge from passages and (ii) structured knowledge from KGs. The strategies applied to collect those type of knowledge is detailed in following sections.

\subsection{Unstructured Knowledge from Passages}
To obtain unstructured knowledge from passages, we consider 
the state-of-the-art pre-trained dense retriever Contriever \citep{DBLP:journals/tmlr/IzacardCHRBJG22}. 

\subsection{Structured Knowledge from KGs}
Compared to passages, the knowledge graph triples have several appealing properties for prompting language models. First, the triples are more compact. Indeed, it takes in most cases less than 20 tokens to encode a triple whereas a 100-word passage would necessitate much more tokens. Besides, triples from KG contain less noise. A passage usually contains more information than the required knowledge to answer a question and would mislead LLMs. %

\subsubsection{KG Triples Retrieval} 
\label{sec:kg_ret}
\begin{figure}
    \centering
    \begin{subfigure}{0.48\textwidth}
        \centering
        \includegraphics[width=0.8\textwidth]{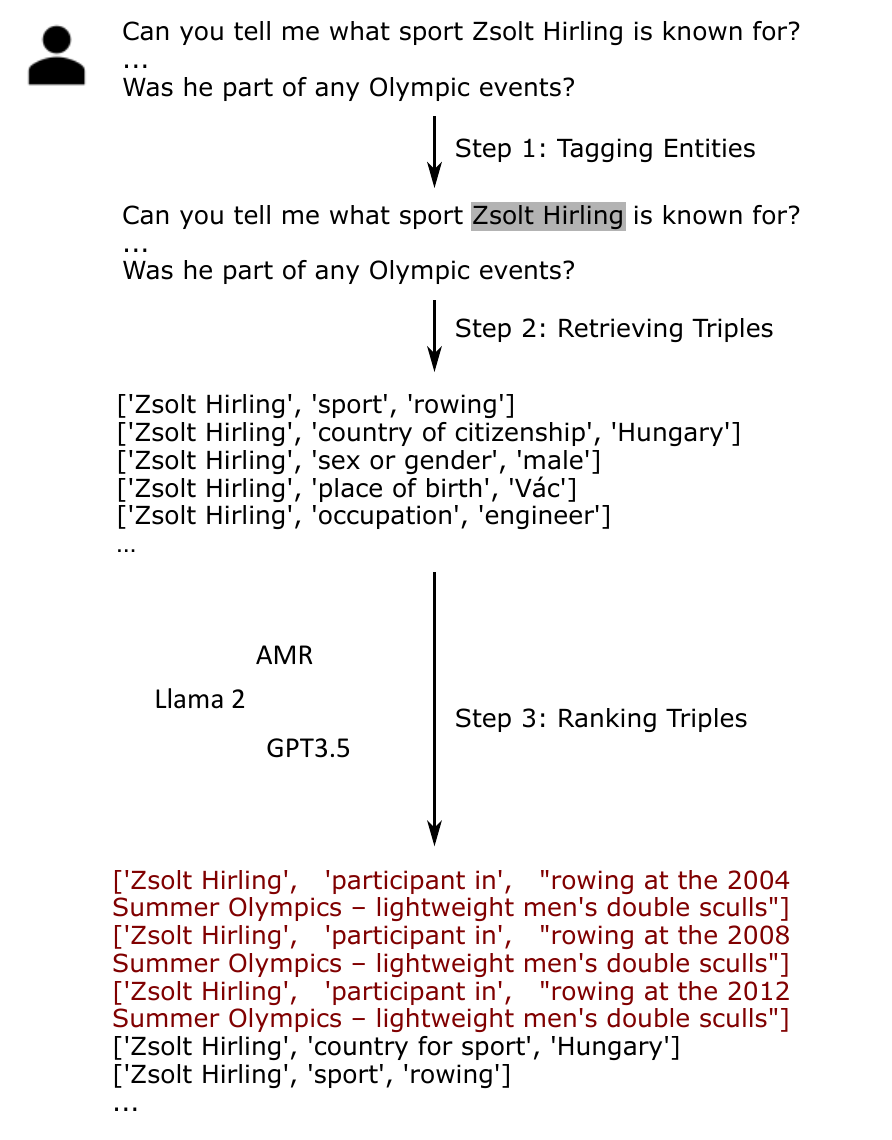}
        \caption{KG Triples Retrieval Pipeline}
        \label{fig:pipeline}
    \end{subfigure}
    \hfill
    \begin{subfigure}{0.48\textwidth}
        \centering
        \includegraphics[width=\textwidth]{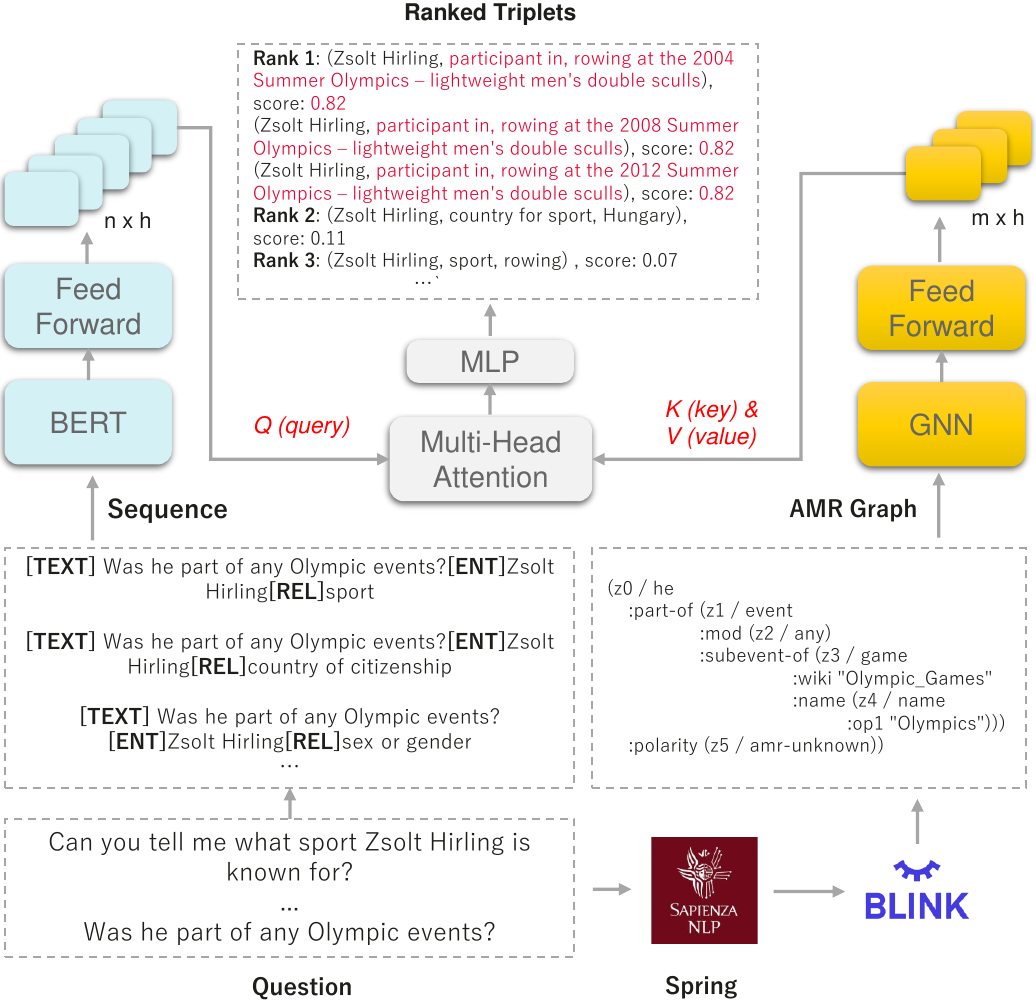}
        \caption{AMR relation linking model}
        \label{fig:amrmodel}
    \end{subfigure}
    \caption{Obtaining non-parametric knowledge from KG.}
    \label{fig:enter-label}
\end{figure}

In order to retrieve pertinent triples from KG, we adopt the commonly used modular KG triples retrieval pipeline from \citep{wang-etal-2021-retrieval} with different triple-ranking baselines. 
As depicted in Figure~\ref{fig:pipeline}, the overall architecture of the KG triples retrieval pipeline is divided into three main components detailed hereinafter.
\paragraph{Step 1: Tagging Entities} 
This primary steps consists in identifying the entities in the input query and map them to an existing KG. To that end, we utilize the TAGME API \citep{DBLP:conf/cikm/FerraginaS10} to align word spans from the questions to Wikidata entities. 
We further filter tagged entities with a threshold of 0.22 (following \citet{mallen-etal-2023-trust}) for the $\rho$ value computed by TAGME API to serve as the tagged entities for the subsequent steps. If there are more than 5 filtered entities for a query, we only keep the top-5 entities based on the $\rho$ value.
\paragraph{Step 2: Retrieving Triples} %
Given the extracted Wikidata entities, we design SPARQL queries to be executed on the Wikidata endpoint\footnote{\url{https://www.wikidata.org/wiki/Wikidata:SPARQL_query_service}} to retrieve the relevant KG triples.
\paragraph{Step 3: Ranking Triples} %
While previous step has successfully returned numerous triples, their usefulness toward resolving the query is questionable. Indeed, several triples are unrelated or out-of-order with respect to their relevance to the query. Thus, a ranking of the triples is performed to select the most promising triples to answer the question. 
We especially view the triples ranking task as a specific instance of the relation linking task.  We explore two different strategies including prompting LLMs and an Abstract Meaning Representation (AMR)-based method. Regarding the former prompting approach, we feed the LLMs with the user query, the tagged entities and the relations from the retrieved triples. We then instruct the model to rank the relations in the prompt according to their relevance in answering the given query. Therefore, LLMs generate a textual output which we parse to extract the rank for each relation. The associated prompt can be found in \ref{app:prompt}. 
In a second phase, we further investigate semantic-based approach to better capture the semantics of the relations thanks to AMR which has proven to benefit knowledge base question answering \citep{kapanipathi-etal-2021-leveraging}. We provide details and elaborate more on AMR approach in Section~\ref{sec:amr}.

\subsubsection{AMR-Based Triples Ranking}
\label{sec:amr}

The AMR graph is a semantic abstract representation of natural language \citep{banarescu-etal-2013-abstract}. This graph effectively captures the semantic relations in sentences by abstracting away from the syntax and has been proven to be helpful in relation linking tasks \citep{naseem-etal-2021-semantics}. The nodes of AMR represent instances, literals, or concepts extracted from the original sentence. These nodes are derived from either framesets from PropBank \citep{kingsbury-palmer-2002-treebank}, normalized surface forms, or specific concepts from the AMR vocabulary, including special entity types. On the contrary, the edges stand for the relations or roles between these nodes which are labelled based on the corresponding frame or from predefined relations from AMR set. Thus, a key component which makes AMR extremely powerful is the fine-grained level of relations between concepts.

We propose to use such a representation to enhance our triple ranking strategy. We frame the ranking task as a binary classification. More specifically, given an entity from the question and a relation from one of the retrieved triples, we aim at predicting whether such a pair is necessary to answer the question or not. We adopt a dual-encoder architecture made of a BERT-based entity-relation pair encoder and a GNN-based AMR encoder. After the fusion of their representation to compute attention scores, a MLP is applied to rank triples. Our proposed architecture consisting of several components is depicted in Figure~\ref{fig:amrmodel} and further detailed below.

\paragraph{Entity-Relation Pairs Encoder}
Given a query $q$, tagged entity $e_q^{(i)} \in E_q$\footnote{$E_q$ is the identified entity list from Step 1 described in Section~\ref{sec:kg_ret}.}, and candidate relation $r_q^{(i)}\in R_q$\footnote{$R_q$ is the candidate relation list from Step 2 described in Section~\ref{sec:kg_ret}, which is the relation list extracted from all retrieved triples.}, we define the Entity-Relation Pairs as '[TEXT]$q$[ENT]$e_q^{(i)}$[REL]$r_q^{(i)}$'. [TEXT], [ENT], and [REL] are special tokens inserted following \citet{naseem-etal-2021-semantics} to identify the input query ($q$), tagged entities $e_q^{(i)}$, and candidate relations $r_q^{(i)}$ respectively. Then we encode each candidate entity-relation pair with BERT \citep{devlin-etal-2019-bert}.

\paragraph{AMR Encoder}
Following \citet{naseem-etal-2021-semantics}, we first generate the AMR graph of the input question using SPRING \citep{DBLP:conf/aaai/BevilacquaBN21} and link the annotated entities with BLINK \citep{wu-etal-2020-scalable}. In order to leverage the rich semantic and graph structure of AMR, we employ a GNN-based encoder. However, since relying on the message passing scheme, GNNs only learn nodes embeddings and thus information carried by the labelled edges would be lost. To this end, we transform our labelled relations as nodes (i.e. reification) so that all information of AMR remains. We adopt the Graph Attention Network (GAT) \citep{DBLP:journals/corr/abs-1710-10903} architecture to learn representations for each node. We expect such embeddings to encompass rich contextual and semantic information.

\paragraph{AMR-Enhanced Multihead-Attention}
Let $E_{AMR} = [e_1, e_2, \ldots, e_n]$ denote the node embeddings resulting from the AMR encoder (GNN), where each $e_i$ is a $d$-dimensional vector corresponding to the $i$-th AMR graph that extracted from a sentence. These embeddings are utilized as both keys ($K$) and values ($V$) in our attention mechanism. Let $Q_{ERP} = [q_1, q_2, \ldots, q_m]$ represent the representations computed by the Entity-Relation Pairs encoder, where each $q_j$ is a $d$-dimensional vector representing the $j$-th entity-relation pair within the sentence, serving as queries ($Q$) in the attention mechanism. Our AMR-Multihead-Attention (AMA) mechanism is defined as follows:
\[
\text{AMA}(Q, K, V) = \text{Concat}(\text{head}_1, \text{head}_2, \ldots, \text{head}_h)W^O
\]

where each attention head, $\text{head}_i$, is computed by:
\[
\text{head}_i = \text{Attention}(QW^Q_i, KW^K_i, VW^V_i)
\]

The attention function is defined as:
\[
\text{Attention}(Q, K, V) = \text{softmax}\left(\frac{QK^T}{\sqrt{d_k}}\right)V
\]

In the equations above:

\begin{itemize}
    \item $W^Q_i$, $W^K_i$, and $W^V_i$ are the parameter matrices for the $i$-th attention head for queries, keys, and values, respectively.
    \item $W^O$ is a parameter matrix for linearly transforming the concatenated outputs of all heads.
    \item $d_k$ is the dimensionality of the key vectors, which is used to scale the dot products in the softmax function, preventing excessively large values that could impede gradient stability.
\end{itemize}

This formulation highlights the dependencies between the AMR and each entity-relation pair. This setup allows for a more dynamic interpretation of entity relations, explicitly influenced by their underlying semantic structures within the sentences.

\section{Experimental Setup}
We evaluate several LLMs on the LTGen benchmark in different settings in order to answer the proposed research questions. Following \citet{DBLP:journals/corr/abs-2310-11511}, we formalise both tasks in the LTGen benchmark as zero-shot generation tasks.
Hyper-parameters and detailed experimental settings can be found in \ref{app:setup}.

\subsection{Large Language Models}
To achieve zero-shot generation, the selected language models must possess the capability to follow instructions. We consider both closed-source LLMs and open-source LLMs.
We select GPT-3.5\footnote{We use \texttt{gpt-3.5-turbo-1106} as it is the latest version at writing.} as the proprietary large-scale LLM by calling the OpenAI API. We choose LLaMA2 \citep{DBLP:journals/corr/abs-2307-09288} as our base open-source model.
We particularly use the chat fine-tuned version. Furthermore, to probe the significiance of the model size on performance, we employ the 7B, 13B, and 70B versions of LLaMA2. 
We utilize the vLLM library \citep{DBLP:conf/sosp/KwonLZ0ZY0ZS23} for fast inference for all selected open-source LLMs.
In addition, we investigate LLM-based triple ranking with GPT-3.5 and LLaMA2.
\subsection{Knowledge Sources}
\paragraph{Passages}
To retrieve passages from external sources, we consider 
the neural-based Contriever models. 
We choose to use the Wikipedia corpus provided by the KILT benchmark\footnote{\url{http://dl.fbaipublicfiles.com/KILT/kilt_knowledgesource.json}} to make sure that the Wikipedia pages of the entities used for creating the LTGen benchmark are included in the passage corpus. Each passage has a maximum length of 100 words. We 
make use of the MSMARCO fine-tuned contriever checkpoint\footnote{\url{https://huggingface.co/facebook/contriever-msmarco}} for better retrieval quality. We retrieve the top-10 passages for each sample of both LTGen-QA and LTGen-Conv.

\paragraph{Knowledge Graphs}
To collect the KG triples, we follow the retrieval pipeline as described in Section~\ref{sec:kg_ret}.
In addition to the TAGME API usage to identify potential topic entities, we also designed an oracle setup where the golden topic entities are directly used. For each method, we use triples with top-5 relations for every topic entity for subsequently prompting language models. For prompt efficiency, we randomly select 10 triples within each entity-relation pair.
\subsection{Prompt Settings}
We formalise both tasks (LTGen-QA and LTGen-Conv) in the LTGen benchmark as zero-shot generation tasks. For all LLMs, we format the prompt as a ``system'' prompt with a following ``user prompt''. The system prompt is an instruction stating to the model the current corresponding task with its respective constraints. Table~\ref{tab:system_prompt} illustrates the system prompts which we found to be effective for all the LLMs in both tasks. The user prompt is the input of both external non-parametric knowledge and the dialogue or question.

\begin{table}[h]
    \centering
    \begin{tabular}{p{2cm}p{10cm}}
    \toprule
        External Knowledge & System Prompt \\ \midrule
        No & You are a helpful assistant. People will provide you with a dialogue, and your task is to respond to the last query in the conversation. \\  \midrule
        Yes & You are a helpful assistant. People will provide you with a dialogue, and your task is to respond to the last query in the conversation. Relevant knowledge that may assist you is provided above the dialogue. \\ \bottomrule
    \end{tabular}
    \caption{System prompts used in this work.}
    \label{tab:system_prompt}
\end{table}

\subsection{Metrics}
We hereinafter introduce our evaluation metrics to assess the performance of the models.
\paragraph{KG Retrieval Metrics}
We measure the performance for KG triples retrieval using recall which is commonly used in information retrieval tasks. We compute recall at relations as we formalise the triple ranking task as relation linking task. For computing the score, we divide the number of correct entity-relation pair with the number of reference entity-relation pairs.

\paragraph{End-to-end Metrics}
Following \citet{DBLP:journals/corr/abs-2310-11511}, we measure three different knowledge matching scores for both LTGen-QA and LTGen-Conv tasks: knowledge matching (KM), exact knowledge matching (eKM) and ratio knowledge matching (rKM). Given the generated prediction $y$ and the reference answer entities in the reference triples $\hat{e}\in\hat{\mathbf{e}}$, these metrics are computed as follows:
\begin{equation}
    KM = \left\{\begin{aligned}
    0, & \exists \hat{e}\in\hat{\mathbf{e}}, \text{str}(\hat{e})\in y \\
    1, & \forall \hat{e}\in\hat{\mathbf{e}}, \text{str}(\hat{e})\notin y
    \end{aligned}
    \right.
\end{equation}
\begin{equation}
    eKM = \left\{\begin{aligned}
    0, & \forall \hat{e}\in\hat{\mathbf{e}}, \text{str}(\hat{e})\in y \\
    1, & \exists \hat{e}\in\hat{\mathbf{e}}, \text{str}(\hat{e})\notin y
    \end{aligned}
    \right.
\end{equation}
\begin{equation}
    rKM = \frac{\text{count}(\hat{e}\in\hat{\mathbf{e}}, \text{str}(\hat{e})\in y)}{|\hat{\mathbf{e}}|}
\end{equation}

In addition to these knowledge matching scores, we also compute the reference response based metrics for the LTGen-Conv dataset. However, similarly to text summarization \citep{DBLP:journals/corr/abs-2209-12356}, we find that traditional overlap-based metrics, such as BLEU \citep{papineni-etal-2002-bleu} and ROUGE \citep{lin-2004-rouge} are not ideal for evaluating performance in tasks involving long-form texts. Previous works \citep{chen-etal-2021-nli-models, DBLP:journals/corr/abs-2208-07316} have shown that Natural Language Inference (NLI) models can serve as a robust evaluator, exhibiting high correlation with human judgements for natural language generation tasks. Thus, we follow \citet{DBLP:journals/corr/abs-2208-07316} and use entailment scores to measure the generation quality of different methods. For a predicted generation, we use an entailment model to measure whether a model's generated content can entail the reference response. There are three scores in this measurement: the entailment score (E), the natural score,
and the contradiction score (C). An optimal prediction should have a high entailment score and a low contradiction score. We report E-C as the overall score. For the choice of entailment model, we use DeBERTa V3\footnote{\url{https://huggingface.co/MoritzLaurer/DeBERTa-v3-large-mnli-fever-anli-ling-wanli}} \citep{DBLP:journals/corr/abs-2111-09543} which has been pre-trained with multiple NLI datasets. 
\section{Results and Analysis}
Table~\ref{tab:overall_res} shows the overall evaluation results of different knowledge settings with different LLMs on the LTGen benchmark. We use the AMR ranking as it outperforms ranking with LLMs (cf. Table~\ref{tab:kg_res}). 
Figure~\ref{fig:res_lt} shows the results with respect to the long-tail level, while Figure~\ref{fig:res_nums} shows the results with respect to reference triple numbers on the LTGen benchmark. 

\subsection{How LLMs Perform without External Non-parametric Knowledge?}
\paragraph{Overall Performance}
We witness that GPT-3.5 works best in the LTGen-QA dataset while in the LTGen-Conv dataset, we observe that Llama 2 70B exhibit better knowledge match scores. Interestingly, when measuring the NLI score, we find that GPT-3.5 beats Llama 2 70B in the same dataset. Though GPT-3.5 only achieves a slightly higher E score compared to Llama 2 70B (0.499 vs 0.483), the much lower C score (0.190 vs 0.267) contributes most to the gap of E-C score between GPT-3.5 and Llama 2.

When looking at the generated examples, we observe that compared to Llama 2 models, GPT-3.5 seems to somehow demonstrate a self-awareness of its own knowledge. Indeed, instead of generating an hallucinated response, GPT-3.5
occasionally refuses to return a proper answer by apologizing (e.g. returning ``Sorry, I can't answer.''). 
\begin{table}[h]
    \centering
    \small
    \begin{tabular}{ccccccccccc}
    \toprule
        \multirow{2}{*}{Model} & \multirow{2}{*}{Params} & \multicolumn{3}{c}{LTGen-QA} & \multicolumn{6}{c}{LTGen-Conv} \\
        & & KM & eKM & rKM & KM & eKM & rKM & E & C & E-C \\\midrule
        \multicolumn{11}{c}{\textit{No Knowledge}}
        \\ \midrule
        Llama 2 & 7B & 0.315 & 0.271 & 0.231 & 0.380 & 0.342 & 0.312 & 0.394 & 0.277 & 0.117 \\
        Llama 2 & 13B & 0.359 & 0.311 & 0.270 & 0.425 & 0.383 & 0.350 & 0.452 & 0.282 & 0.170 \\
        Llama 2 & 70B & 0.393 & 0.345 & 0.302 & \underline{0.497} & \underline{0.452} & \underline{0.415} & 0.483 & 0.267 & 0.215 \\
        GPT-3.5 & UNK & \underline{0.405} & \underline{0.359} & \underline{0.318} & 0.488 & 0.442 & 0.404 & \underline{0.499} & \underline{0.190} & \underline{0.309} \\ \midrule
        \multicolumn{11}{c}{\textit{Passages (Contriever)}}
        \\ \midrule
        Llama 2 & 7B & 0.478 & 0.427 & 0.381 & 0.450 & 0.406 & 0.370 & 0.501 & 0.168 & 0.334 \\
        Llama 2 & 13B & 0.496 & 0.446 & 0.401 & 0.436 & 0.394 & 0.358 & 0.493 & 0.152 & 0.340 \\
        Llama 2 & 70B & 0.511 & 0.462 & 0.417 & 0.567 & 0.520 & 0.479 & 0.607 & 0.175 & 0.432 \\
        GPT-3.5 & UNK & 0.510 & 0.462 & 0.418 & 0.541 & 0.490 & 0.448 & 0.575 & 0.148 & 0.427 \\\midrule
        \multicolumn{11}{c}{\textit{KG Triples (AMR)}}
        \\ \midrule
        Llama 2 & 7B & 0.743 & 0.685 & 0.634 & 0.600 & 0.554 & 0.511 & 0.560 & 0.136 & 0.425 \\
        Llama 2 & 13B & 0.745 & 0.692 & 0.643 & 0.602 & 0.562 & 0.522 & 0.594 & 0.125 & 0.469 \\
        Llama 2 & 70B & 0.750 & 0.708 & 0.669 & 0.732 & 0.692 & 0.652 & 0.677 & 0.124 & 0.553 \\
        GPT-3.5 & UNK & 0.771 & 0.732 & 0.696 & 0.728 & 0.687 & 0.648 & 0.692 & 0.096 & 0.596 \\\midrule
        \multicolumn{11}{c}{\textit{Passages + KG Triples}}
        \\ \midrule
        Llama 2 & 7B & 0.738 & 0.681 & 0.630 & 0.606 & 0.560 & 0.516 & 0.608 & 0.104 & 0.504 \\
        Llama 2 & 13B & 0.740 & 0.690 & 0.643 & 0.589 & 0.546 & 0.506 & 0.618 & 0.098 & 0.519 \\
        Llama 2 & 70B & 0.750 & 0.707 & 0.664 & 0.729 & 0.686 & 0.643 & 0.721 & 0.104 & 0.617 \\
        GPT-3.5 & UNK & 0.756 & 0.710 & 0.667 & 0.717 & 0.671 & 0.626 & 0.705 & 0.092 & 0.613 \\
        \bottomrule
        
    \end{tabular}
    \caption{Overall end-to-end evaluation results.}
    \label{tab:overall_res}
\end{table}

\begin{figure}[t]
    \centering
    \includegraphics[width=\linewidth]{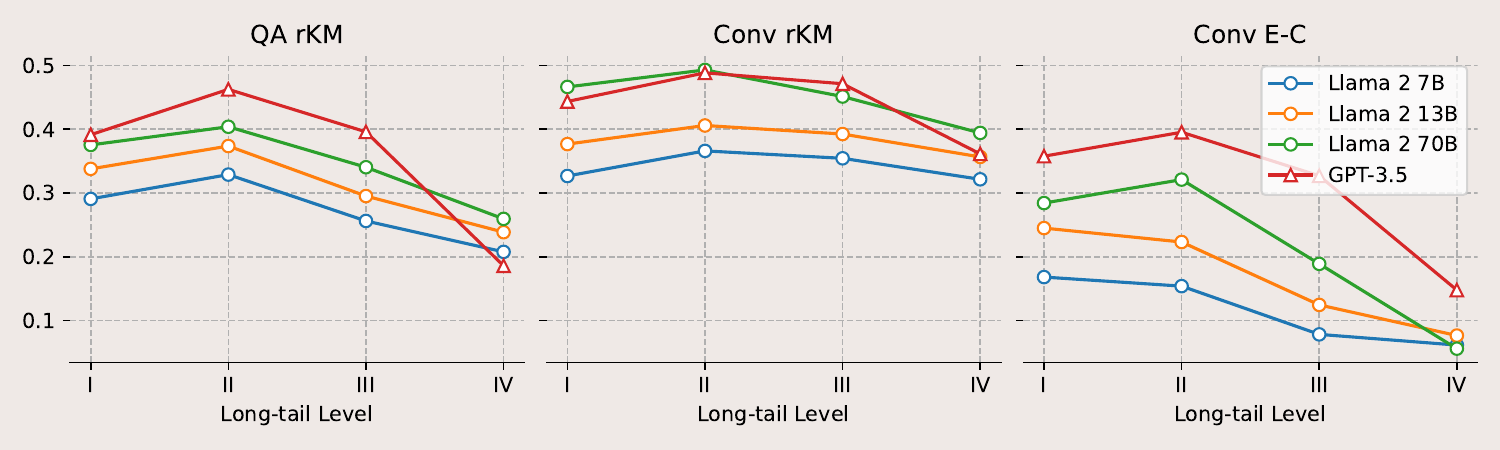}
    \caption{LLMs' performance on the LTGen benchmark with respect to long-tail level.}
    \label{fig:res_lt}
\end{figure}

\begin{figure}[t]
    \centering
    \includegraphics[width=\linewidth]{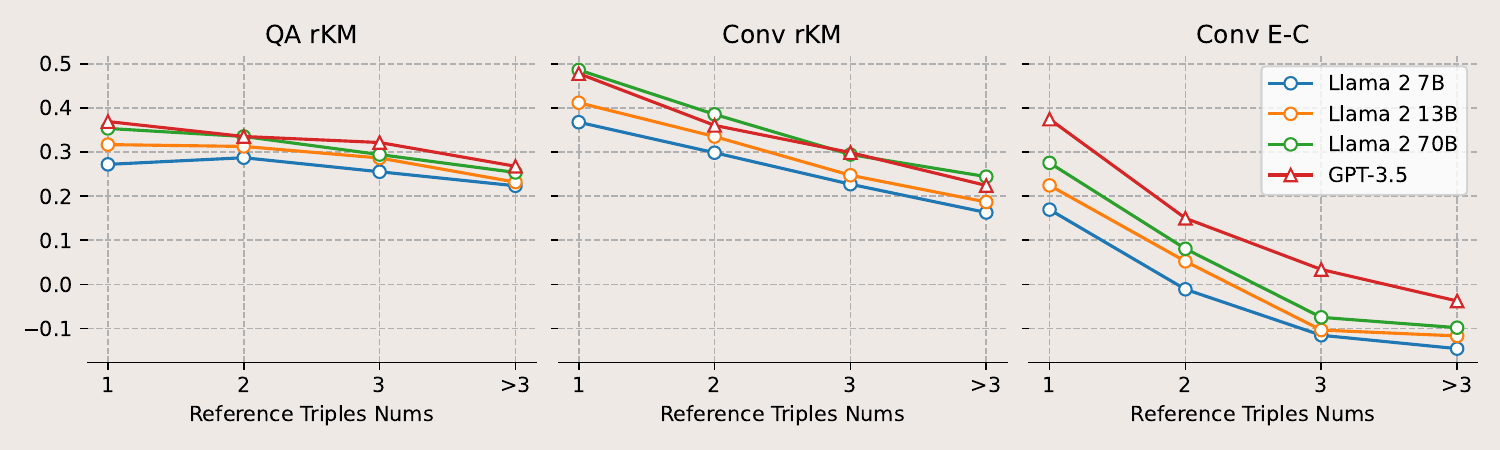}
    \caption{LLMs' performance on the LTGen benchmark with respect to reference triple numbers.}
    \label{fig:res_nums}
\end{figure}

\paragraph{Long-tail Level}

Unsurprisingly, we observe significant drops in performance with respect to the long-tail level. The higher the rarity level, the worst the performance (Figure~\ref{fig:res_lt}). This is in line with our intuition that LLMs cannot well-handle long-tail facts. Interestingly, we however remark a quite consistent performance increase from the long-tail level I to level II. We consider the diversity in relations from level I and therefore the need for more reference triples to be the culprit of a light difficulty increase in that level\footnote{The average reference triples numbers for long-tail level I and II are 1.38 and 1.30 respectively.}. In addition to that discovery., we also observed that GPT-3.5 achieves the lowest rKM score on the LTGen-QA dataset because it generates more unsure responses compared to Llama 2 models. This can be proved in the LTGen-Conv results, where GPT-3.5 achieves the best E-C score but not the highest rKM score, indicating that GPT-3.5 is more robust to prevent hallucination when the long-tail level is high.

\paragraph{Reference Triple Numbers}

From Figure~\ref{fig:res_nums}, we can observe that more reference triples result in a harder task for LLMs. When considering the E-C score, even the best GPT-3.5 model gets a negative value when there are more than three reference triples.

\begin{table}[!h]
    \centering
    \begin{tabular}{cccccc}
    \toprule
        \multirow{2}{*}{Model} & \multirow{2}{*}{Params} & \multicolumn{2}{c}{LTGen-QA}& \multicolumn{2}{c}{LTGen-Conv} \\
         & & Recall & Recall (oracle) & Recall & Recall (oracle) \\ \midrule
        \multicolumn{6}{c}{\textit{LLM-based relation ranking}} \\ \midrule
        Llama 2 & 7B & 0.587 & 0.840 & 0.471 & 0.719 \\
        Llama 2 & 13B & 0.588 & 0.860 & 0.493 & 0.749 \\
        Llama 2 & 70B & 0.651 & 0.935 & 0.550 & 0.826 \\
        GPT-3.5 & UNK & 0.619 & 0.895 & 0.532 & 0.812 \\ \midrule
        \multicolumn{6}{c}{\textit{AMR-based relation ranking}} \\ \midrule
        AMR & $\sim$0.2B & \textbf{0.672} & \textbf{0.991} & \textbf{0.560} & \textbf{0.846} \\
    \bottomrule
    \end{tabular}
    \caption{KG triples retrieval results for different KG triples ranking methods on LTGen benchmark.}
    \label{tab:kg_res}
\end{table}

\subsection{How Different Ways of Obtaining KG Knowledge Works?}

In this section, we compare different relation-linking approaches applied in our proposed KG triples retrieval pipeline. Table~\ref{tab:kg_res} shows the relation-linking result. Compared with using LLMs, the AMR relation linking approach achieves a higher Recall score in both datasets of the LTGen benchmark with only less than 0.2B parameters. Since there might be accumulative errors from the entity tagging step, we further measure the relation-linking result in an oracle setting where the entity is pretended to be correctly linked. The best AMR relation linking approach received close to perfect Recall on the LTGen-QA dataset and over 84\% Recall on the LTGen-Conv dataset. This indicates that there still is space for significant improvement in prompting with KG triples. A reasonable direction for future work can be to improve the performance of the entity tagging approach to fill the gap between Recall and Recall (oracle). It is worth noting that the relation-linking performance on LTGen-QA is a bit higher than that on the LTGen-Conv dataset. This comes in two reasons: 1) each question on the LTGen-QA dataset is only related to a single relation, making the relation-linking task easier; 2) the context window of the LTGen-Conv dataset is much larger, making both entity tagging and relation linking more challenging tasks.

\subsection{How Different Non-parametric Knowledge Help LLMs?}
\begin{figure}[!h]
    \centering
    \includegraphics[width=\linewidth]{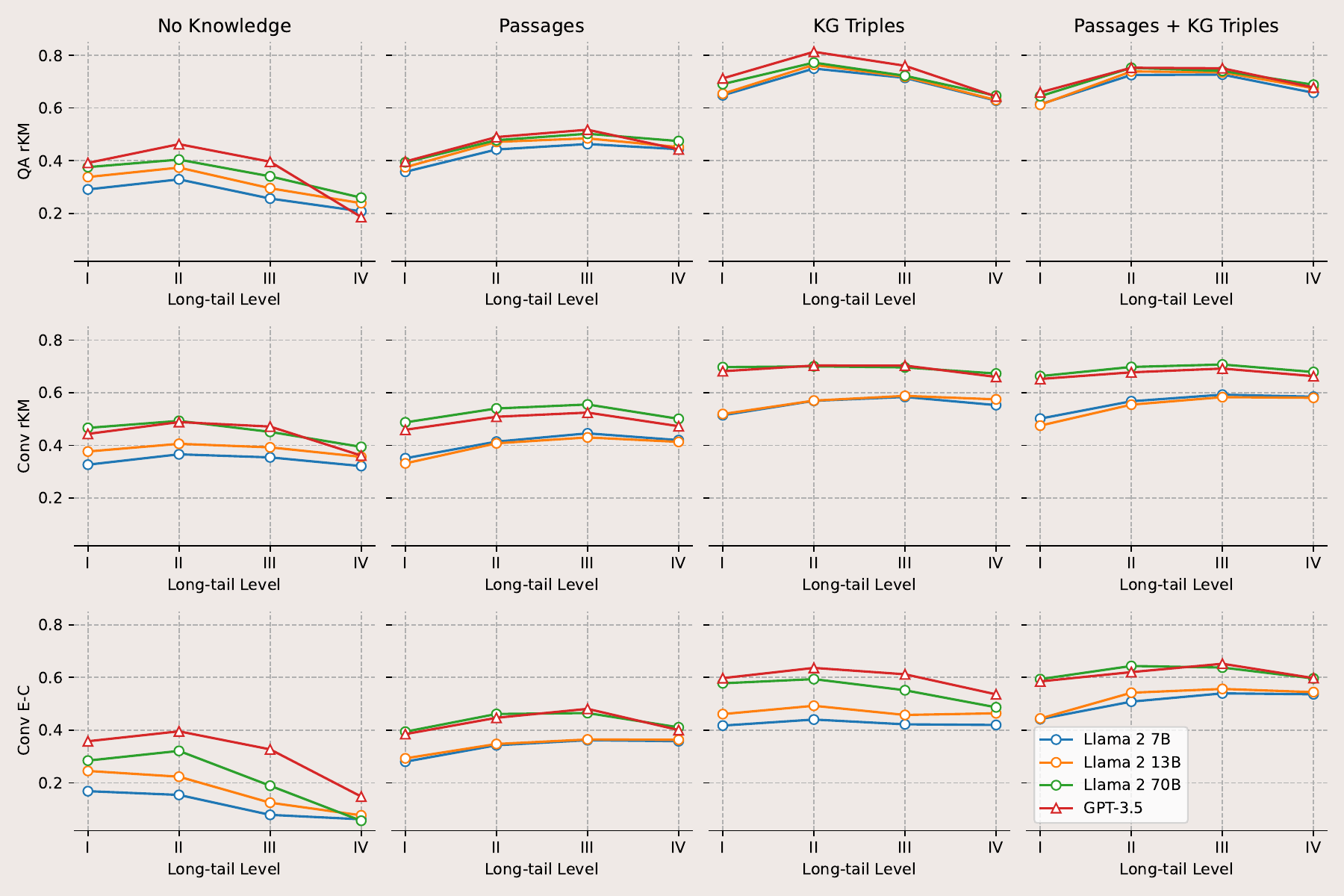}
    \caption{LLMs' performance when prompted with different sources of external knowledge on the LTGen benchmark with respect to long-tail level.}
    \label{fig:res_lt_all}
\end{figure}

\begin{figure}[!h]
    \centering
    \includegraphics[width=\linewidth]{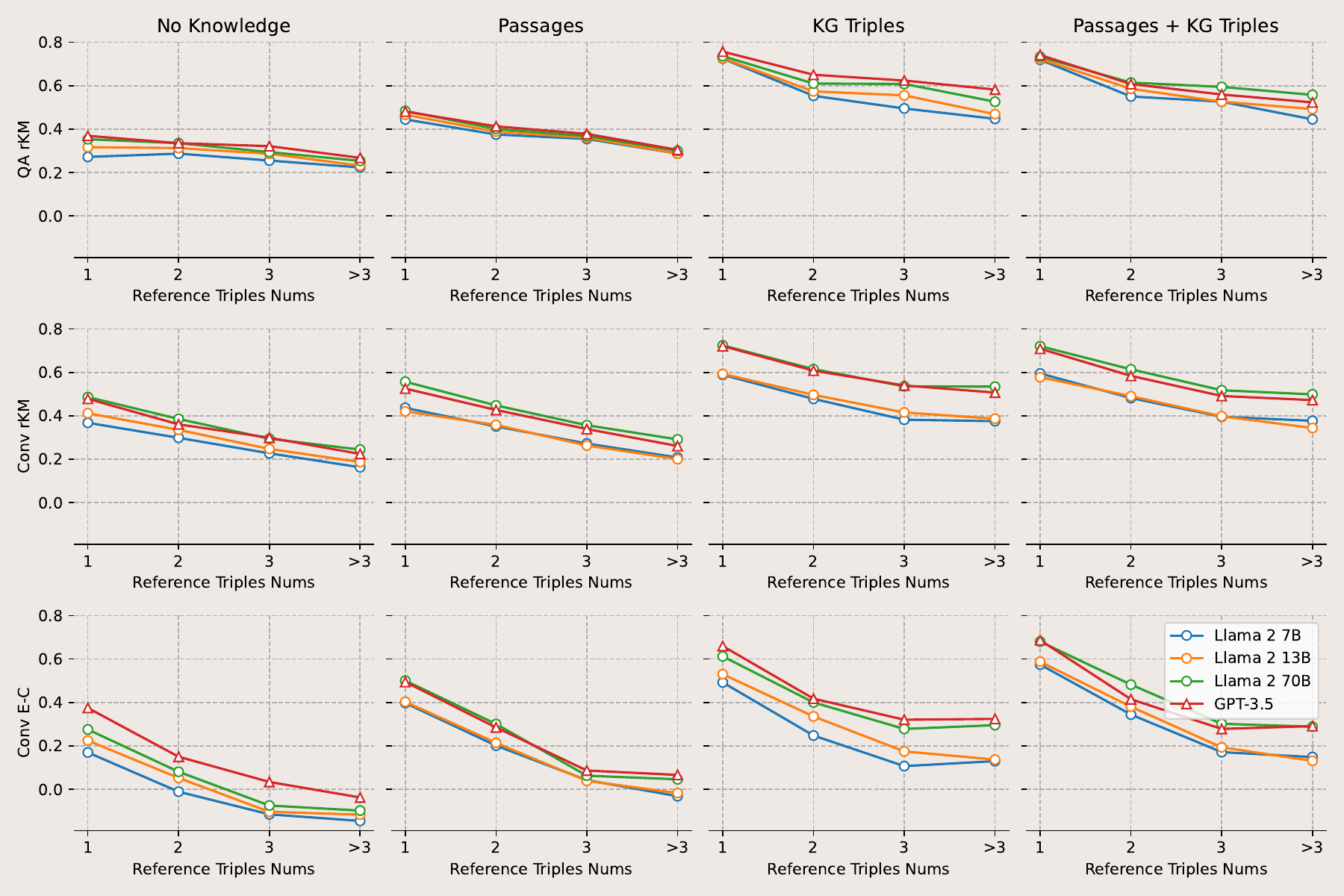}
    \caption{LLMs' performance when prompted with different sources of external knowledge on the LTGen benchmark with respect to reference triple numbers.}
    \label{fig:res_nums_all}
\end{figure}

\paragraph{Overall Performance}
Prompting LLMs with non-parametric knowledge shows a consistent improvement compared with the no knowledge setting (cf. Table~\ref{tab:overall_res}), with %
the improvement of passages is not as significant as that of KG triples. In addition, in the LTGen-QA dataset, smaller models usually benefit more when prompted with external knowledge. Furthermore, we can see a notable decrease in C scores among all models with non-parametric knowledge. This indicates that LLMs benefit from non-parametric knowledge in reducing hallucinations. By prompting LLMs with KG triples, we achieve much better performance in all metrics as opposed to prompting LLMs with passages which further leads to higher inference costs\footnote{The average input context token numbers of Llama 2 is 701.9 by prompting KG triples and 1014.1 by prompting top-5 contriever retrieved passages.}, showing both the effectiveness and efficiency of using structural knowledge from KGs. By merging both passage information and KG triples information, we do not observe a performance improvement upon the knowledge matching metrics. However, the NLI-based score is improved in the LTGen-Conv dataset. We consider that with different sources of external knowledge, LLMs are more cautious in providing responses, thus reducing hallucinations compared to single-sourced prompting. 

\paragraph{Long-tail Level}
From Figure~\ref{fig:res_lt_all}, we can observe that when the long-tail level is low (typically level I and II), prompting LLMs with passages benefit limited or even harm the performance in both datasets. When the long-tail level increases, the performance gain grows as well. These are similar to the observations from \citet{mallen-etal-2023-trust}. As for KG triples, all models perform better with passages across all long-tail levels. It's worth noting that in the LTGen-Conv dataset, there is quite a clear gap between relatively smaller models (7B and 13B) and large-scale models (70B and above). This gap is more clear when prompting external knowledge. This might indicate that smaller models get harder to handle both dialogue history and external knowledge.

\paragraph{Reference Triple Numbers}
Similar to the no knowledge setting, the performance of LLMs generally drops as the number of reference triples increases. However, we can observe that the performance gap between different LLMs prompted with KG triples is more clear than prompted with passages. (cf. Figure~\ref{fig:res_nums}). In general, LLMs with KG triples perform better than with passage. We also note that the performance drop from 3 triples setting to more than 3 triples setting is less sharp when KG triples as part of the external knowledge source, indicating that within limited context length, KG triples is a richer knowledge source that could capture more factual knowledge.

\section{Conclusion and Future Works}
In this paper, we have introduced a novel pipeline for automatically building question-answering and  conversational datasets. We use this pipeline to create the LTGen benchmark, which has two datasets (LTGen-QA and LTGen-Conv) for assessing the ability of LLMs to ground long-tail knowledge. We conducted evaluations on different LLMs with varying types of non-parametric knowledge using our benchmark. 

Our key findings include: (1) non-parametric knowledge helps LLMs to improve their performance, especially when the required knowledge has a high rarity level or is composed of multiple triples. Our hypothesis is motivated by  PopQA~\cite{mallen-etal-2023-trust}, which is based on texts generated from triples via pre-defined templates over a limited  number of relations. Our comprehensive experiments    demonstrate non-parametric knowledge helps LLMs, not only for simple short-form question answering tasks (LTGen-QA) but also for complex free-form conversational question answering tasks (LTGen-Conv); (2) using prompts in the form of knowledge graph triples is more effective than relying on passages; (3) %
 even though prompts from  multiple sources of external knowledge (passages and KG triples) are correct, LLMs might not take them into account when generating their outputs. Thus retrieval based validations seem necessary.   %

Future works could focus on   building safe and robust LLMs, including verifying and reducing grounding error where LLMs do not follow the context information we provide \citep{baek-etal-2023-knowledge-augmented-language}. Furthermore, we observe a huge performance improvement in the KG retrieval pipeline by using golden entities, showing the benefit of effective entity tagging approaches that should be investigated in future works.

\appendix
\section{Results in KILT benchmark}
Table~\ref{tab:kilt_res} shows the evaluation results on two KIG datasets in the KILT benchmark. We can see that GPT-3.5 with only parametric knowledge outperform the state-of-the-art RE2G methods with external non-parametric knowledge on the TQA dataset.
\label{app:kilt}
\begin{table}[h]
    \centering
    \begin{tabular}{cccc}
    \toprule
        Model & Dataset & EM & F1 \\ \midrule
        GPT3.5 & NQ & 36.06 & 50.07 \\
        GPT3.5 & NQ-LT & 34.77 & 45.17 \\ \midrule
        GPT3.5 & TQA & \textbf{79.57} & \textbf{86.54} \\
        GPT3.5 & TQA-LT & 77.90 & 85.41 \\ \midrule\midrule
        RE2G & NQ & \textbf{46.70} & \textbf{62.44} \\
        RE2G & TQA & 74.01 & 80.86 \\
    \bottomrule
    \end{tabular}
    \caption{GPT3.5 and RE2G \citep{glass-etal-2022-re2g} performance in NQ and TQA development set. RE2G is the state-of-the-art retrieval-augmented generation method.}
    \label{tab:kilt_res}
\end{table}

\section{Prompts}
\label{app:prompt}
We detail the prompts used for the creation of both LTGen-QA and LTGen-Conv benchmark datasets as described in Section~\ref{sec:benchmark_construction}. The variable of the prompt is highlighted in {\color{red}red}.

Prompts for LTGen-QA:
\begin{quote}
    \color{blue}Create a question for each subject entity and relation pairs. The question should be in natural language and like what human user will ask. Your response format should be as follows: 

    Entity Relation Pair: $<$entity relation pair chosen$>$ 
    
    Question: $<$generated question$>$ 
    
    ... 
    
    Entity Relation Pairs: 
    
    \color{red}{[relation pairs]}

\end{quote}

Prompts for LTGen-Conv:
\begin{quote}
    \color{blue}Create a natural conversation with an agent and a user. The user ask consistent questions that require external knowledge from the triples listed below (MUST be natural and consistent, using prepositions, if not sure about the proper prepositions, the user should first ask for related questions (or the agent should first provide the related information)). The agent responds to the user and selects triples from the list below to be the ONLY knowledge source. Attach the chosen triples (if applicable) as well. The user NEVER asks questions that can be answered without the selected triples! Your response format should be as follows: 
    
    User: {\color{red}$<$question$>$}
    
    Agent: {\color{red}$<$answer$>$} 
    
    T: $<$selected triples$>$
    
    ... 
    
    Knowledge Graph Triples: 
    
    \color{red}{[triples]}

\end{quote}

For utilizing LLMs for relation linking, we use the following prompt for both GPT3.5 and Llama 2 as below:
\begin{quote}
    system prompt:
    
    {\color{blue}You are a helpful assistant. People will provide you with a question, an entity and several relations, and your task is to rank the relation based on the relevance for answering the question. Only response top-5 relations without any other things, one for each line.}
    
    user prompt:
    
    \color{blue}{Question: {\color{red}$<$question$>$}
    
    Entity: {\color{red}$<$entity$>$}
    
    Relations: {\color{red}$<$relations$>$}}
\end{quote}

\section{Data Quality}
\label{app:data_quality}

We engage two annotators to review the associated triple annotations for 400 data samples from LTGen-Conv dataset. The selected samples from uniformly sampled from each long-tail level. We ask both annotators to follow the instruction below to deciding whether a dialogue is correctly annotated by GPT:

\begin{quote}
The annotation of reference triples is correct if and only if: 

1) the annotated reference triples fully support the query and answer pair.

2) the answer requires no more knowledge than the reference triples.

3) there is no unused triples annotated.
\end{quote}

Below is an example which violate the above three rules and is recognised by both annotators as incorrectly annotated sample where the reference triples can not support the reasoning from query to answer (dialogue history is omitted):
\begin{itemize}
    \item Query: Is Otis Clay an American artist?
What religion did Ayoub Tabet practice?
    \item Reference Output: 
Yes, "Tryin To Live My Life Without You" is from the United States of America which aligns with Otis Clay's nationality.
    \item Listed Reference Triples:
    \color{red}{("Tryin' To Live My Life Without You", 'country of origin', 'United States of America')}
\end{itemize}

\section{Experiments Details}
\label{app:setup}
\subsection{Environment}
For prompting LLaMa 2 7B and 13B, we use a single A100 80G GPU with vLLM library \citep{DBLP:conf/sosp/KwonLZ0ZY0ZS23}. For prompting LLaMA 2 70B, we use two A100 80G GPUs with vLLM library. For training and inferencing AMR-based relation linking model, we use one RTX 4090 24G GPU.
\subsection{Hyperparameters}
Table~\ref{tab:hparams} shows the hyperparameters setup for the prompting experiments. For all un-mentioned hyper-parameters, we simply set them to their default values.
\begin{table}[h]
    \centering
    \small
    \begin{tabular}{c|cc}
    \toprule
        Hyper-parameter & GPT-3.5 & LLaMA 2 \\
         \midrule
        temperature & 0. & 0. \\
        max new tokens & 128 & 128\\
        \bottomrule
    \end{tabular}
    \caption{Hyper-parameters setup for the experiment.}
    \label{tab:hparams}
\end{table}

\bibliographystyle{elsarticle-num-names} 
\bibliography{references, anthology}

\end{document}